\documentclass[letterpaper]{article} 
\usepackage[preprint]{aaai2027}  
\usepackage[hyphens]{url}  
\usepackage{graphicx} 
\usepackage{natbib}  
\usepackage{caption} 
\usepackage{algorithm}
\usepackage{algorithmic}
\usepackage{amsmath}
\usepackage{amssymb}
\usepackage{newfloat}
\usepackage{listings}

\usepackage{colortbl} 
\definecolor{chronorow}{gray}{0.9}
\newcommand{\sourcenote}{}  

\newcommand{\iconWhere}{\raisebox{-0.1ex}{\scalebox{1.05}{$\blacktriangledown$}}}
\newcommand{\iconWhen}{\raisebox{-0.1ex}{\scalebox{1.05}{$\circlearrowright$}}}

\DeclareCaptionStyle{ruled}{labelfont=normalfont,labelsep=colon,strut=off} 
\floatstyle{ruled}
\newfloat{listing}{tb}{lst}{}
\floatname{listing}{Listing}

\usepackage{booktabs}
\usepackage{nameref}

\title{Chronosphere: Space-Time Tessellation of Local Climate Experts}
\author{
    Daniel Cher\textsuperscript{\rm 1},
    Eric Xing\textsuperscript{\rm 1},
    Kexing Li\textsuperscript{\rm 1},
    Brian Wei\textsuperscript{\rm 1},
    Isaac Corley\textsuperscript{\rm 2},
    Nathan Jacobs\textsuperscript{\rm 1}
}
\affiliations{
    \textsuperscript{\rm 1}Washington University in St. Louis\\
    \textsuperscript{\rm 2}Taylor Geospatial Institute\\
    \{cher, e.xing, ariana.l, b.j.wei, jacobsn\}@wustl.edu,
    isaac.corley@taylorgeospatial.org
}

\makeatletter
\let\Chrono@maketitle\@maketitle
\def\@maketitle{%
  \Chrono@maketitle
  \vskip 0.5em
\begin{center}
\includegraphics[width=0.63\textwidth]{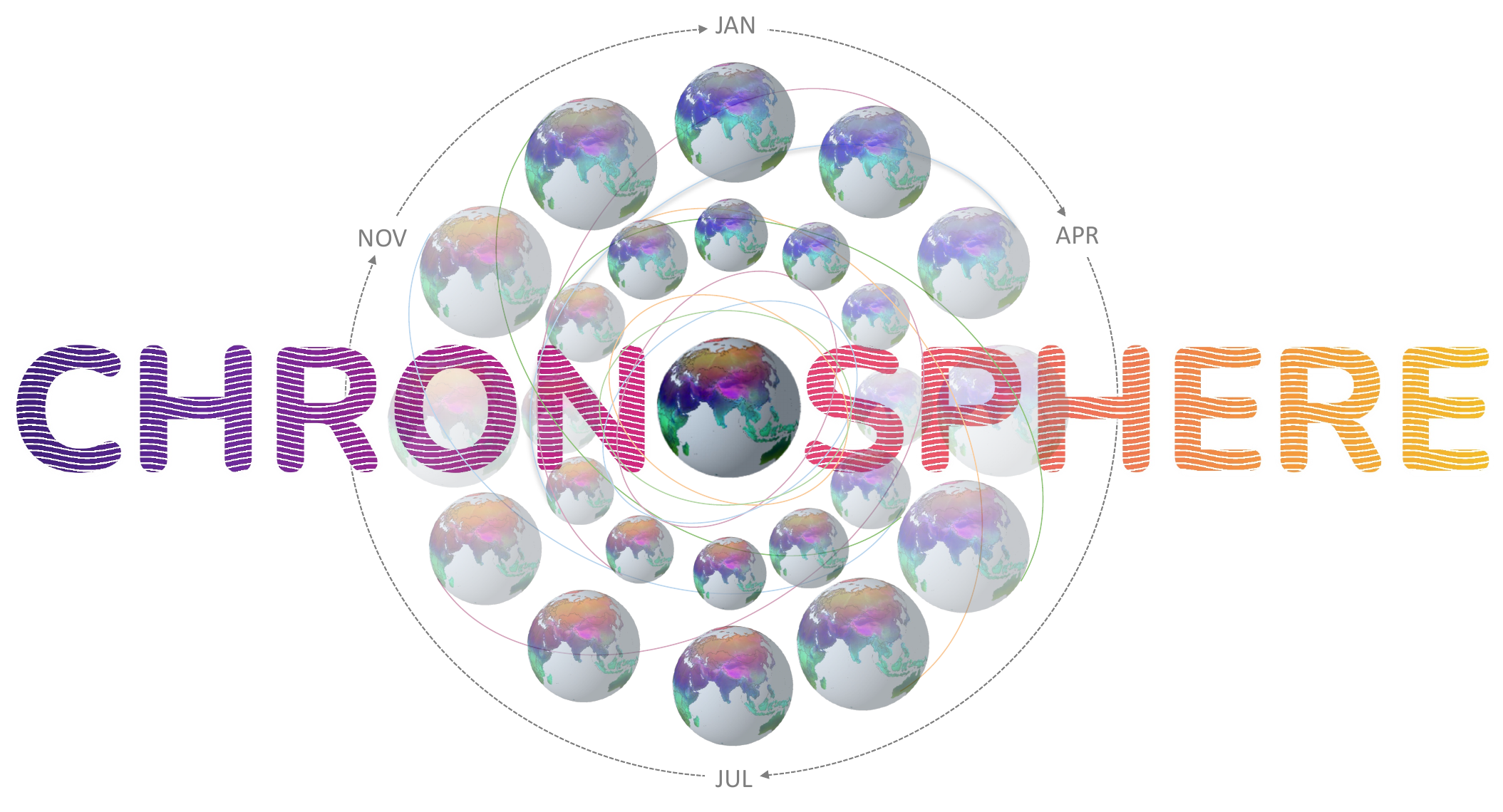}
\captionof{figure}{\textbf{Chronosphere is a single learned spatio-temporal representation of Earth's climate.} Each globe shows the frozen embedding queried at a different point in the year (Jan $\to$ Dec clockwise from the top), projected to RGB over land. The embedding shifts smoothly from month to month, so one representation captures both where a place sits and how it changes through the year.}
\label{fig:teaser}
\end{center}
\vskip 1em

}
\makeatother

\begin{document}

\maketitle

\begin{abstract}

We introduce \emph{Chronosphere}, a spatio-temporal neural field that learns representations of climate. A central challenge in geographic representation learning is modeling environmental processes whose spatial and temporal complexity varies widely. Yet existing location encoders typically fix a single level of detail everywhere. Global bases such as spherical harmonics spread capacity uniformly across space and time. Localized bases resolve only predefined regions. Learned tessellations adapt, but are inefficient at representing higher frequencies. \emph{Chronosphere} unifies these approaches, pairing an adaptive tessellation of learnable sites on the spacetime torus $S^2\times S^1$ with a shared bank of local basis functions. Both where capacity is placed and how much detail each region carries adapt to the data, across space and time. Trained to reconstruct climatology, \emph{Chronosphere} matches or leads state-of-the-art location encoders across spatial and temporal tasks, with the largest gains under spatial and temporal transfer. \sourcenote

\end{abstract}

\section{Introduction}
\label{sec:intro}

\begin{figure*}[!t]
\centering
\includegraphics[width=0.8\textwidth]{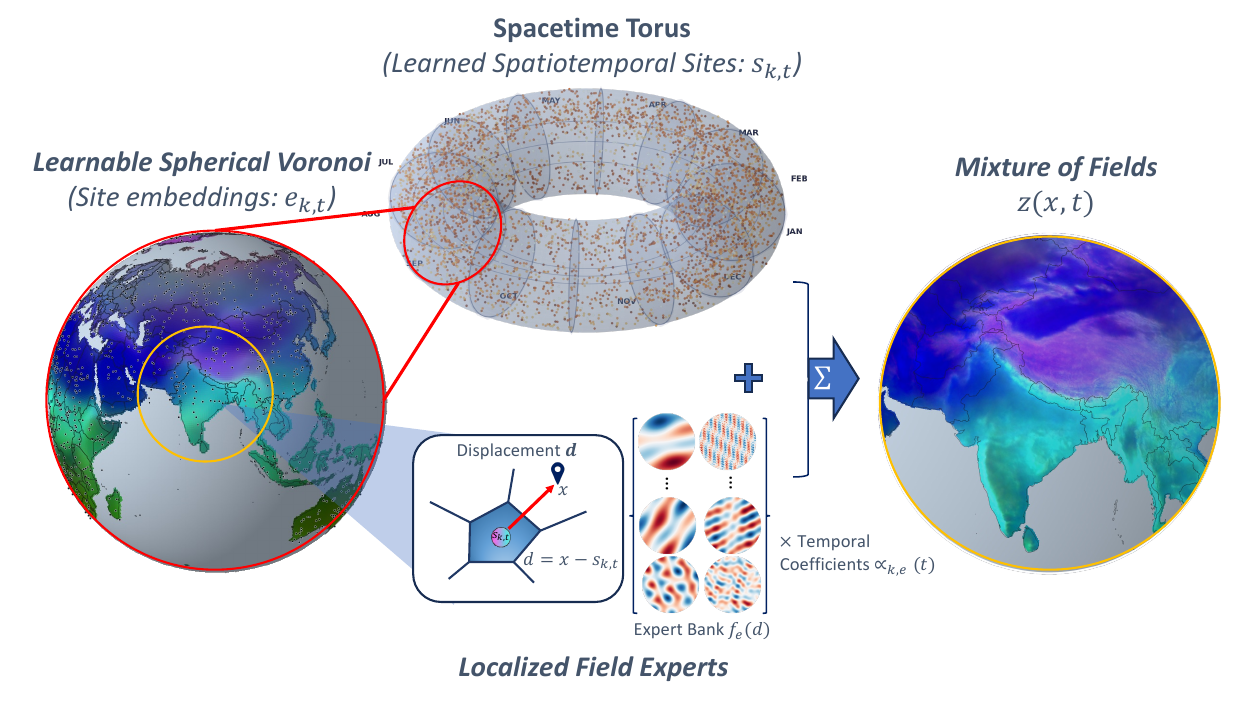}
\caption{\textbf{Chronosphere.} A query, a location $x\in S^2$ and within-year time $t$, is softly assigned to learnable sites on the spacetime torus $S^2\times S^1$. At each site a shared bank of sine experts $f_j$ is mixed by per-site coefficients $\alpha_{k,j}(t)$ and added to the aggregated site embedding to form the location vector $z(x,t)$, which the decoder reconstructs into climate fields.}
\label{fig:mixture_fields}
\end{figure*}

Location encoders have become a popular tool in geospatial modeling, turning a coordinate into a reusable representation of the environment. They have proven useful across tasks such as species distribution modeling~\cite{cole2023sinr,mac2019presence}, crop-yield estimation~\cite{tseng2025galileo}, air-quality forecasting~\cite{karimzadeh2025pm25}, canopy-height mapping~\cite{lang2023canopy}, and satellite image synthesis~\cite{sastry2024geosynth,cher2026vectorsynth,wei2026terraditomega}. Much of this work has focused on well-sampled settings, where a probe fit on the representation interpolates among nearby labels. But these representations matter most where local labels are scarce, when a task must generalize to a distant region or an unseen season~\cite{cai2025nolocation}. That need is sharpening as climate conditions migrate across the landscape~\cite{loarie2009velocity,milly2008stationarity} and the timing of the seasons shifts with them~\cite{burrows2011pace,williams2007novel}, carrying the environment away from where any model was trained. A useful representation must therefore generalize across space and time, which depends on capturing how the environment itself varies.

Environmental fields vary in frequency across both space and season~\cite{rolf2024mission}. Temperature varies little across the Amazon basin but changes sharply over the neighboring Andes. Likewise, the Sahel remains nearly uniform for much of the year until the onset of the monsoon produces a strong moisture gradient within a matter of weeks. A representation should therefore concentrate resolution where the field is sharp and stay coarse where it is smooth, rather than fixing the level of detail across the globe and year. Existing encoders fall short of this in different ways. Global bases such as spherical harmonics cover the sphere but spread capacity uniformly~\cite{russwurm2024siren}, localized bases such as Slepian functions concentrate resolution only within predefined regions~\cite{rao2026localized}, and learned tessellations let sites adapt to the data but gain finer structure only by adding more sites~\cite{cher2026tte}, never enriching each one. Temporal signals enter as uniform global conditioning applied identically at every location~\cite{dollinger2025climplicit,mickisch2025ste}. These approaches have been pursued in isolation, but their strengths are complementary. Combining the parameter efficiency of a shared basis with the adaptivity of a learned tessellation, without predefined regions, gives a representation that generalizes where labels are sparse.

We introduce \emph{Chronosphere} (Fig.~\ref{fig:teaser}), a spatio-temporal neural field built on this idea (Fig.~\ref{fig:mixture_fields}). Its learnable sites live on the spacetime torus $S^2 \times S^1$, each carrying a spatial and temporal position, and queries are encoded as soft Voronoi weightings over joint spatio-temporal distance. Each site draws on a shared bank of local basis-function experts spanning a range of frequencies, mixing them differently through the year to set how finely its region is resolved and how that detail shifts with the seasons. Trained to reconstruct climatology, the frozen embedding learns fine-grained spatial variation across the globe (Fig.~\ref{fig:mixture_fields}) and seasonal variation that matches the geography of the annual cycle (Fig.~\ref{fig:seasonal}).

Because these representations matter most where labels are scarce, we evaluate under both interpolation and transfer, using spatial block and temporal holdouts. \emph{Chronosphere} matches or leads existing geospatial and spatiotemporal encoders across a range of spatial and temporal tasks, with its largest advantage on interpolation of temporal tasks and under spatial or temporal transfer, where local autocorrelation is minimal. 

The main contributions of this work are:
\begin{itemize}
    \item \emph{Chronosphere}, a spatio-temporal neural field that unifies a learnable Voronoi tessellation with a bank of local basis functions, adapting capacity and detail across space and time.
    \item An evaluation across a suite of spatial and temporal tasks, where \emph{Chronosphere} matches or beats the best encoders in-distribution and leads them under generalization to unseen regions and seasons.
    \item Qualitative and quantitative evidence that the learned representation matches real climate structure, from fine-grained spatial fields to the phase and geography of the yearly cycle.
\end{itemize}

\section{Related Work}
\label{sec:related_work}

\subsection{Spatio-Temporal Representation Learning}

Location encoders differ most in the signal that shapes their embeddings~\cite{mai2022review}. Many align coordinates with street-level~\cite{cepeda2023geoclip,yin2019gps2vec}, satellite~\cite{klemmer2025satclip,cher2026tte}, and citizen-science~\cite{mai2023csp} imagery through contrastive learning, while SINR~\cite{cole2023sinr} instead trains on species occurrence, and RANGE~\cite{dhakal2025range} adds retrieval augmentation to recover the fine spatial detail contrastive pretraining tends to lose. These encoders yield useful representations of places, but each maps a location to a single spatial embedding and none capture how that location changes through the year.

Models that do incorporate time typically add it as a global signal. CLIMPLICIT~\cite{dollinger2025climplicit} pretrains a location encoder to regress monthly CHELSA climatology, so that its embedding is organized by climate and carries a seasonal signal. That signal enters as a single global function conditioned on the month, rather than as structure that varies from place to place. Joint space-time encoders~\cite{mickisch2025ste,Shatwell_2025_ICCV} learn location and time together in a single network, but still fix how finely each place is resolved rather than adapting it across space and season. Chronosphere shares CLIMPLICIT's climate-field objective but builds season into the site geometry itself, giving the representation its own spatio-temporal structure.

\subsection{Adaptive Capacity Allocation in Neural Fields}

Most encoders allocate representational capacity through a global basis over the whole sphere, whether multi-scale sinusoids~\cite{mai2020space2vec,mai2023sphere2vec} or spherical harmonics with a SIREN~\cite{russwurm2024siren}, spreading it uniformly. Slepian bases~\cite{rao2026localized} instead concentrate resolution within regions fixed in advance. 

Learnable partitions make this allocation adaptive. Spherical Voronoi tessellations and related decompositions~\cite{Di_Sario_2026_CVPR,rebain2021derf,cher2026tte} learn where sites are placed so that capacity follows the data, yet each site holds a single smooth vector with fixed within-region detail. Local implicit fields~\cite{sitzmann2020siren,tancik2020fourier} supply that detail, but as one global function with no partition to place it. Each gives either an adaptive partition or rich within-region detail, never both. Chronosphere unifies them, pairing a learnable partition with a bank of local basis-function experts so that both where capacity sits and how finely each region resolves adapt to the data, jointly across space and season.

\section{Data}
\label{sec:data}

Chronosphere uses two complementary climate corpora, monthly CHELSA climatology (${\sim}$1\,km, sharp spatial structure) and daily ERA5-Land reanalysis (${\sim}$9\,km, within-year and across-year dynamics). The monthly variant trains on CHELSA alone, while the daily variant trains jointly on both. Downstream evaluation datasets are described in the \nameref{sec:experimental_setup} section (full provenance tables are deferred to the appendix).

\paragraph{Monthly climatology (CHELSA).}
\label{sec:data:chelsa}
We train the monthly CHELSA branch on v2.1 monthly climatologies (WMO 1981--2010 normal) at ${\sim}$1\,km over land~\cite{karger2017chelsa}. Because this is a climatological normal, the corpus emphasizes spatial structure and monthly seasonality without year-to-year variation. Following CLIMPLICIT~\cite{dollinger2025climplicit}, the model regresses eleven normalized variables spanning temperature, precipitation, radiation, wind, humidity, and derived moisture indices, listed in full in the appendix (Training Data Variables).

\paragraph{Daily reanalysis (ERA5-Land).}
\label{sec:data:era5}
The ERA5 branch uses daily sequences (1970--2024) for within-year and year-resolved modeling~\cite{munozsabater2021era5land}. It covers the same land climate variables as CHELSA, minus cloud cover, but at a coarser ${\sim}$9\,km native resolution that limits fine-scale spatial detail.

\paragraph{Climate indices (inter-annual drivers).}
\label{sec:data:indices}
Daily reanalysis captures seasonal structure, but year-to-year variation (slow forced trends and shorter-lived anomalies) requires an explicit inter-annual signal. A free per-year embedding can memorize in-sample years but does not define years out of distribution. We condition the daily model on eight monthly climate indices (ENSO, NAO, PDO, AMO, global mean temperature, log CO$_2$, stratospheric aerosol, and sunspot activity): large-scale drivers whose reconstructions extend to the mid-19th century, so the same conditioning can apply in training, backcast, and forecast once index values are supplied. The indices are inputs, not reconstruction targets. We describe how they modulate the predictions in the \nameref{sec:methodology:interannual} subsection.

\begin{table*}[!t]\centering
\caption{\textbf{Downstream performance} (frozen encoders). Each cell is iid\,/\,extrapolation; \textbf{bold} = best, \textit{italic} = second best, ranked separately within each half. Scores are means over three seeds. Chronosphere's per-task standard deviation stays under $0.01$ on every iid split. Under extrapolation it stays below ${\sim}0.03$ on most tasks, rising to $0.04$--$0.06$ on the few most sensitive to the split (plant traits, snow, elevation, streamflow).
}
\label{tab:main}
\scriptsize
\setlength{\tabcolsep}{4pt}
\begin{tabular}{lccccc|ccccc}
\toprule
& \multicolumn{5}{c|}{\textbf{Spatial} \;(iid\,/\,$10^\circ$ geo-extrap.)} & \multicolumn{5}{c}{\textbf{Within-year} \;(iid\,/\,window-extrap.)} \\
\cmidrule(lr){2-6}\cmidrule(lr){7-11}
& \multicolumn{2}{c}{Acc.\,$\uparrow$} & \multicolumn{2}{c}{$R^2\,\uparrow$} & \multicolumn{1}{c|}{RMSE\,$\downarrow$} & \multicolumn{3}{c}{$R^2\,\uparrow$} & AUC\,$\uparrow$ & AP\,$\uparrow$ \\
\cmidrule(lr){2-3}\cmidrule(lr){4-5}\cmidrule(lr){6-6}\cmidrule(lr){7-9}\cmidrule(lr){10-10}\cmidrule(lr){11-11}
Model & Bio & Eco & Elev & Traits & Canopy & NDVI & Snow & Streamflow & Fire & USA-NPN \\
\midrule
SINR        & 0.67/0.55 & 0.48/0.12 & 0.57/$-$0.05 & 0.59/0.12 & 6.1/7.3 & 0.68/0.59 & 0.48/$-$0.30 & 0.08/$-$0.38 & 0.81/0.79 & 0.11/0.09 \\
CSP-iNat    & 0.59/0.46 & 0.54/0.08 & 0.34/0.13 & 0.42/0.20 & 7.0/8.0 & 0.56/0.47 & 0.44/$-$0.33 & 0.08/$-$0.37 & 0.77/0.76 & 0.10/0.09 \\
GeoCLIP     & 0.60/0.41 & 0.60/0.09 & 0.39/$-$0.06 & 0.37/0.15 & 8.2/9.6 & 0.41/0.40 & 0.39/$-$0.31 & 0.05/$-$0.36 & 0.75/0.75 & 0.11/0.10 \\
SatCLIP     & 0.69/0.54 & 0.69/0.20 & 0.66/0.32 & 0.50/0.31 & 5.9/6.3 & 0.69/0.60 & 0.45/$-$0.29 & 0.10/$-$0.38 & 0.82/0.80 & 0.11/0.09 \\
TTE         & 0.77/0.59 & 0.67/0.28 & 0.84/0.65 & \textbf{0.67}/\textit{0.41} & 5.1/5.6 & 0.76/0.66 & 0.50/$-$0.28 & 0.10/$-$0.38 & \textit{0.84}/0.82 & 0.11/0.10 \\
\midrule
GT-Loc      & 0.72/0.59 & 0.71/0.14 & 0.62/0.31 & 0.51/0.39 & 5.8/6.3 & 0.71/0.67 & 0.57/0.03 & 0.21/$-$0.13 & 0.82/0.81 & 0.20/\textit{0.11} \\
STE & \textbf{0.83}/0.58 & 0.75/\textbf{0.32} & 0.88/0.69 & 0.62/0.17 & 5.0/5.9 & \textit{0.79}/0.77 & 0.78/0.68 & 0.50/0.21 & \textit{0.84}/\textit{0.84} & \textit{0.23}/\textit{0.11} \\
CLIMPLICIT  & \textit{0.81}/0.58 & \textbf{0.79}/0.28 & 0.89/0.75 & 0.57/0.30 & 5.1/5.6 & 0.77/0.73 & 0.81/0.70 & 0.42/0.12 & 0.81/0.78 & 0.22/0.09 \\
\midrule
\rowcolor{chronorow} Chronosphere (monthly) & \textbf{0.83}/\textit{0.65} & \textit{0.78}/\textit{0.31} & \textbf{0.94}/\textbf{0.87} & \textbf{0.67}/\textbf{0.47} & \textbf{4.5}/\textit{5.1} & \textbf{0.87}/\textit{0.84} & \textit{0.85}/\textit{0.76} & \textit{0.53}/\textit{0.31} & \textbf{0.88}/\textbf{0.87} & \textit{0.23}/\textbf{0.12} \\
\rowcolor{chronorow} Chronosphere (daily)   & \textit{0.81}/\textbf{0.66} & 0.76/0.28 & \textit{0.92}/\textit{0.78} & \textit{0.66}/\textbf{0.47} & \textit{4.7}/\textbf{5.0} & \textbf{0.87}/\textbf{0.86} & \textbf{0.89}/\textbf{0.85} & \textbf{0.58}/\textbf{0.40} & \textbf{0.88}/\textbf{0.87} & \textbf{0.25}/\textbf{0.12} \\
\bottomrule
\end{tabular}
\end{table*}

\begin{figure*}[!t]
\centering
\includegraphics[width=0.30\textwidth]{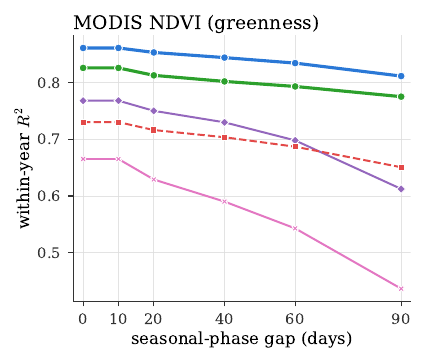}\hfill
\includegraphics[width=0.30\textwidth]{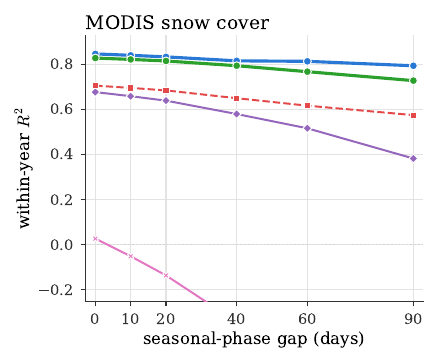}\hfill
\includegraphics[width=0.30\textwidth]{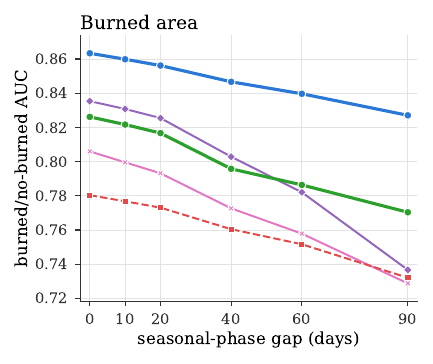}\\[4pt]
\includegraphics[width=0.75\textwidth]{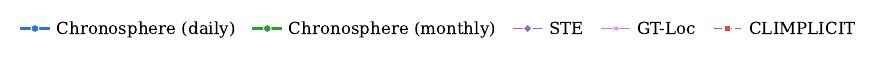}
\caption{\textbf{Temporal extrapolation across seasonal phase.} Probe skill as the seasonal-phase gap $\delta$ between the training periods and the held-out window grows, from in-distribution (left) to a full season held out (right). Greenness and snow are scored by $R^2$ and burned area by burned/no-burned ROC-AUC, with the dotted line marking chance. Time-aware encoders only.}
\label{fig:temporal_extrap}
\end{figure*}

\section{Experimental Setup}
\label{sec:experimental_setup}

\paragraph{Downstream tasks.}
We evaluate frozen encoders on downstream tasks chosen for a strong climate-driven signal, grouped by the temporal structure they require (Table~\ref{tab:downstream_tasks}).
\emph{Spatial niche} tasks (WHERE) ask what kind of place a coordinate represents: biome, ecoregion, and elevation from the RANGE benchmark~\cite{dhakal2025range}, plant functional traits~\cite{lusk2026traits}, and canopy height~\cite{lang2023canopy}.
\emph{Within-year} tasks (WHEN) probe the seasonal cycle via MODIS NDVI~\cite{didan2015mod13}, snow cover~\cite{hall2016mod10}, river discharge~\cite{kratzert2023caravan}, GFED burned area~\cite{vanderwerf2017gfed}, and USA-NPN phenology~\cite{usanpn,rosemartin2018npn}.

\begin{center}
\small
\setlength{\tabcolsep}{4pt}
\captionof{table}{Climate-driven downstream tasks (\iconWhere\ WHERE, \iconWhen\ WHEN).}
\label{tab:downstream_tasks}
\begin{tabular*}{\columnwidth}{@{\extracolsep{\fill}} c l l l @{}}
\toprule
 & Task & Target & Resolution \\
\midrule
\iconWhere & Biome & Biome label & Static \\
\iconWhere & Ecoregion & Ecoregion label & Static \\
\iconWhere & Elevation & Elevation & Static \\
\iconWhere & Plant traits & SLA, LDMC, LNC, LPC & Static \\
\iconWhere & Canopy height & Canopy height (m) & Static \\
\addlinespace[2pt]
\iconWhen & MODIS NDVI & Greenness & 16-day \\
\iconWhen & MODIS snow & Snow cover & 8-day \\
\iconWhen & Discharge & Streamflow & Daily \\
\iconWhen & GFED fire & Burned area & 8-day \\
\iconWhen & USA-NPN & Phenophases & Visit date \\
\bottomrule
\end{tabular*}
\end{center}

\vspace{0.3em}

\paragraph{Evaluation protocol.}
For temporal tasks we evaluate frozen encoders at each task's native resolution. We use the embedding at the matching day, month, or year and never pool across time. Spatial niche tasks assign one feature vector to each location. We run the encoder at twelve monthly or at twenty-four day-of-year steps for temporal models, then concatenate the mean, standard deviation, minimum, and maximum so the linear probe captures local seasonality, while retaining a consistent feature dimension across models and tasks. We follow the previous testing protocols~\cite{dhakal2025range,klemmer2025satclip}. Every task is scored with a cross-validated ridge probe, ridge classification for the categorical targets and ridge regression for the continuous ones, so all encoders are compared under one fixed protocol. The appendix (RANGE Linear Protocol) gives the full protocol.

Our main table reports interpolation on three random 80/20 splits and spatial extrapolation on held-out $10^\circ$ lat/lon tiles. We also report results at tile sizes of $5^\circ$ $\to$ $80^\circ$ to trace how accuracy changes with geographic separation. A random split places test points inside the spatial and temporal autocorrelation range of the training data, so interpolation scores largely reflect memorized local context and can overstate how far an encoder transfers~\cite{roberts2017crossvalidation,kattenborn2022spatial,ploton2020spatial}. Blocked holdouts remove that leakage. Separating whole tiles in space and contiguous windows in season measures the representation's value where autocorrelation gives nothing to lean on, the regime that matters as the climate moves into unseen states~\cite{valavi2019blockcv}.
Within-year evaluations hold out contiguous blocks of the seasonal cycle.

\section{Methodology}
\label{sec:methodology}

Chronosphere maps a unit-sphere coordinate $x \in S^2$ and a within-year time $t$ to a site embedding for downstream probing (Fig.~\ref{fig:mixture_fields}). It lets the representation adapt capacity and detail where necessary across space and time. We train two variants of the same architecture. The monthly model reconstructs CHELSA climatologies alone at ${\sim}$1\,km ($P{=}12$). The daily model jointly reconstructs CHELSA and ERA5-Land ($P{=}365$). Our flagship model uses $K{=}8192$ sites and $E{=}48$ experts over six frequency bands.

\paragraph{Spatiotemporal sites.}
Each site behaves like a weather station pinned to both a place and a time of year, and a query, itself a place $x$ and a day $t$, is described by a soft blend of the stations nearest it in both. We place $K$ learnable sites, each with a direction $s_k \in S^2$, a day-of-year $t_k$, a spatial temperature $\tau_k$, a temporal weight $\beta_k$, and an embedding $e_k \in \mathbb{R}^{D}$, so each site is a point on the spacetime torus $S^2 \times S^1$. A single softmax assigns the query to sites and averages their embeddings into a location vector $z_0(x,t)$,
\begin{equation}
\begin{split}
w_k(x,t) &= \mathrm{softmax}_k\big(
\underbrace{\tau_k\,(x \cdot s_k)}_{\text{spatial closeness}}
\;-\;
\underbrace{\beta_k\,\delta_k(t)}_{\text{temporal distance}}
\big), \\
z_0(x,t) &= \sum_{k \in K} w_k(x,t)\, e_k ,
\end{split}
\label{eq:spacetime}
\end{equation}
where $\delta_k(t) = \tfrac{1}{2}\big(1 - \cos(2\pi(t - t_k)/P)\big)$ measures cyclic distance in the year and is zero when the query is the site's day $t_k$.

Temporal resolution then emerges as spatial resolution does. Spatial detail comes from many sites at different directions competing for a query, and fine seasonal detail from sites that overlap in space but sit at staggered days of the year. At one location a summer-anchored and a winter-anchored site hold different embeddings, and a query's weight crossfades between them as the year turns, continuously across the year boundary. This adds a temporal axis to the soft Voronoi tessellation~\cite{cher2026tte}. We tie the seasonal weight to the spatial scale as $\beta_k = \tau_k\, e^{\gamma_k}$ and initialize it small with the $t_k$ spread across the year, so space dominates at first and the assignment reduces to the purely spatial encoder as $\beta_k \to 0$, adding seasonal structure only where the data supports it. Two simpler ways to add time, passing the day-of-year only to the decoder or evolving site embeddings through a low-rank seasonal flow, are compared in Table~\ref{tab:ablation_temporal}.

\paragraph{Local Field Experts.}
A tessellation alone gives each site a single vector, so sharper structure would require adding sites, which quickly becomes computationally expensive. We instead add local texture as a continuous per-site field from a shared bank of sine experts (Fig.~\ref{fig:mixture_fields}). Each expert $f_j$ is evaluated on a normalized displacement from the site that puts every site on a common unit scale,
\begin{equation}
d_k(x) = \sqrt{\tau_k}\,(x - s_k).
\label{eq:displacement}
\end{equation}
A site then forms its local field as a weighted sum of the experts' outputs $f_j\big(d_k(x)\big)$ (Eq.~\ref{eq:displacement}), with its own coefficients $\alpha_{k,j}(t)$ and a readout $W_j$. These per-site fields are blended by the same soft assignment $w_k$ over a small fixed neighborhood of the query's $m$ nearest sites $\mathcal{T}_m(x)$,
\begin{equation}
r(x,t) = \sum_{k \in \mathcal{T}_m(x)} w_k(x,t)\,
\underbrace{\sum_{j=1}^{E} \alpha_{k,j}(t)\, W_j\, f_j\big(d_k(x)\big)}_{\text{per-site local field}},
\label{eq:field_decoder}
\end{equation}
and the location vector is $z(x,t) = z_0(x,t) + r(x,t)$. The coefficients $\alpha_{k,j}(t)$ vary through the year, so a site's local detail shifts with the season, while the experts $f_j$ are fixed spatial shapes shared across all sites. The readout $W_j$ is zero-initialized, so the residual $r$ starts at zero and the model begins as the plain substrate, earning within-region detail only where the data supports it. The flagship uses sine experts, though the basis is a plug-in choice, with alternatives studied in the appendix (Architectural Ablations).

\begin{figure}
\centering
\includegraphics[width=.95\linewidth]{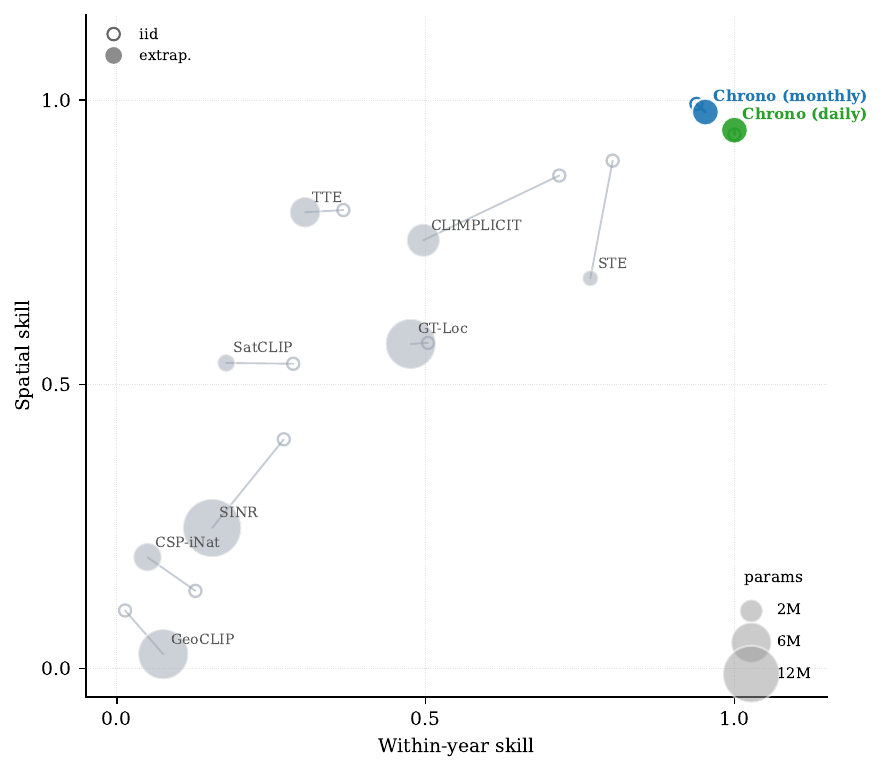}
\caption{\textbf{In-distribution versus out-of-distribution performance.} 
Each model runs from a hollow marker at its in-distribution skill to a filled marker under extrapolation, with marker area the parameter count. A short segment toward the top-right means the encoder generalizes across both axes, and a long segment toward the bottom-left means it collapses out of distribution. Each axis is the min--max-normalized mean across that half's tasks (best model ${\approx}1$, worst ${\approx}0$).
}
\label{fig:skill_bubble}
\end{figure}

\begin{table}\centering
\caption{\textbf{Chronosphere ablations} (within-year, iid\,/\,window-extrap.). Chronosphere (daily) is from Table~\ref{tab:main}. Spatial ablations are in the appendix (Architectural Ablations).}
\label{tab:ablation_temporal}
\scriptsize
\setlength{\tabcolsep}{3pt}
\begin{tabular}{lccccc}
\toprule
setting & NDVI & Snow & Streamflow & Fire & NPN \\
\midrule
\rowcolor{chronorow} Chronosphere (daily) & 0.87/0.86 & 0.89/0.85 & 0.58/0.40 & 0.88/0.86 & 0.25/0.12 \\
\midrule
\multicolumn{6}{l}{\emph{Time in assignment} (vs.\ spacetime)} \\
cyclic & 0.85/0.81 & 0.84/0.73 & 0.57/0.37 & 0.88/0.86 & 0.25/0.12 \\
flow & 0.83/0.81 & 0.83/0.76 & 0.53/0.27 & 0.87/0.85 & 0.23/0.12 \\
\midrule
\multicolumn{6}{l}{\emph{Local field} (vs.\ free mixture)} \\
none (tessellation only) & 0.82/0.79 & 0.86/0.79 & 0.55/0.31 & 0.85/0.81 & 0.23/0.10 \\
gated & 0.87/0.85 & 0.88/0.84 & 0.56/0.34 & 0.87/0.86 & 0.24/0.11 \\
top-$k$ & 0.87/0.84 & 0.87/0.82 & 0.55/0.31 & 0.87/0.86 & 0.24/0.11 \\
\midrule
\multicolumn{6}{l}{\emph{Site count} $K$ (vs.\ $8192$)} \\
$K{=}1024$ & 0.87/0.85 & 0.86/0.79 & 0.55/0.31 & 0.86/0.85 & 0.24/0.12 \\
$K{=}2048$ & 0.86/0.85 & 0.86/0.82 & 0.56/0.32 & 0.87/0.86 & 0.24/0.12 \\
$K{=}4096$ & 0.86/0.85 & 0.86/0.83 & 0.57/0.33 & 0.87/0.86 & 0.24/0.12 \\
$K{=}16384$ & 0.87/0.86 & 0.89/0.85 & 0.57/0.40 & 0.87/0.86 & 0.25/0.12 \\
\midrule
\multicolumn{6}{l}{\emph{Frequency ladder} (vs.\ full bank)} \\
low-freq & 0.86/0.85 & 0.88/0.84 & 0.56/0.38 & 0.87/0.86 & 0.24/0.12 \\
mid-freq & 0.87/0.86 & 0.89/0.85 & 0.57/0.39 & 0.87/0.86 & 0.24/0.12 \\
high-freq & 0.87/0.86 & 0.90/0.86 & 0.58/0.40 & 0.88/0.86 & 0.23/0.11 \\
\bottomrule
\end{tabular}
\end{table}
\begin{figure*}[!t]
\centering
\begin{minipage}[c]{0.58\textwidth}
\centering
\includegraphics[width=\linewidth]{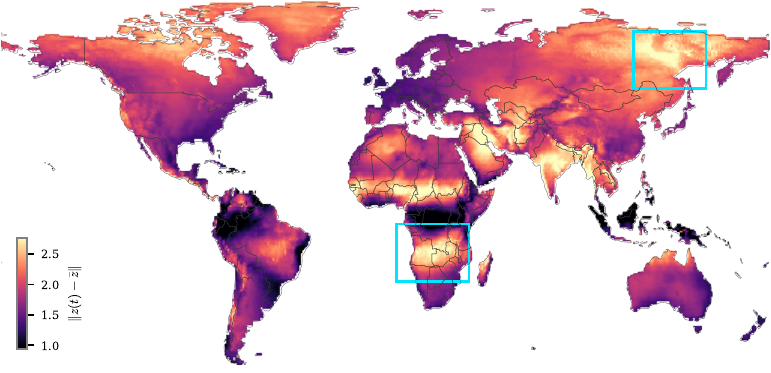}\\[7pt]
{\small Seasonal amplitude $\lVert z(t)-\bar z\rVert$}
\end{minipage}\hfill
\begin{minipage}[c]{0.40\textwidth}
\centering
\makebox[0.33\linewidth]{\footnotesize Jan}\hfill
\makebox[0.33\linewidth]{\footnotesize Apr}\hfill
\makebox[0.33\linewidth]{\footnotesize Jul}\\[1pt]
\includegraphics[width=0.33\linewidth]{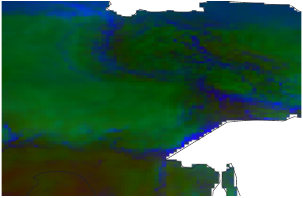}\hfill
\includegraphics[width=0.33\linewidth]{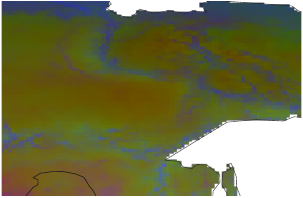}\hfill
\includegraphics[width=0.33\linewidth]{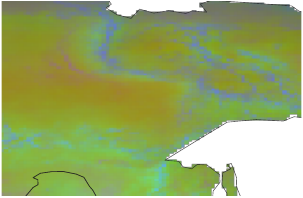}\\[1pt]
{\footnotesize NE Siberia}\\[13pt]
\includegraphics[width=0.33\linewidth]{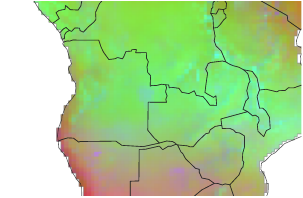}\hfill
\includegraphics[width=0.33\linewidth]{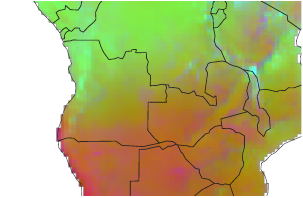}\hfill
\includegraphics[width=0.33\linewidth]{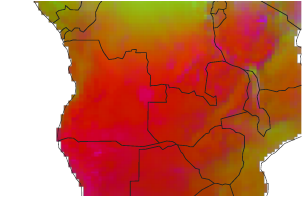}\\[1pt]
{\footnotesize S-Central Africa}
\end{minipage}
\caption{\textbf{The location vector carries seasonal structure that differs in phase and character across the globe.} \emph{Left:} seasonal amplitude of the monthly vector, $\lVert z(t)-\bar z\rVert$, the distance the representation travels from its annual mean, with boxes marking the two regions shown at right. \emph{Right:} the vector (PCA$\,\rightarrow$\,RGB, one shared basis) in January, April, and July for NE Siberia and S-Central Africa. The two regions peak in opposite seasons and move along different color axes, one temperature-driven and one precipitation-driven, so a single shared embedding captures both when each cycle turns and what drives it.}
\label{fig:seasonal}
\end{figure*}

\paragraph{Decoder and embedding.}
Because time is already in $z(x,t)$, the climatology decoder is a two-block residual MLP that reads $z$ alone. Under joint training, the daily variant feeds the shared substrate to two reconstruction branches: a CHELSA head for eleven monthly variables and an ERA5 head for ten daily variables. The monthly variant uses CHELSA alone. We probe using the decoder's frozen $256$-dimensional penultimate activation.

\paragraph{Hybrid inter-annual head.}
\label{sec:methodology:interannual}
To backcast and forecast rather than reconstruct climatology alone, we add a small inter-annual head that puts a year-to-year residual on the climatology, driven by a set of standardized climate indices. The residual is linear in the indices, so raising one index by a fixed amount shifts the reconstruction by that index's learned spatial pattern. The full parameterization and a comparison of alternatives are in the appendix (Yearly Index Drivers).

\paragraph{Training.}
The monthly variant minimizes mean squared error on CHELSA targets over randomly sampled land locations and months.
The daily variant jointly minimizes MSE on ERA5 and CHELSA samples, plus an $\ell_2$ prior that keeps the inter-annual residual small.
Full hyperparameters, optimizer groups, and ablation configurations are in the appendix (Training Details).
\section{Results and Discussion}
\label{sec:results_discussion}

\begin{figure*}[!t]
\centering
\begin{minipage}[t]{0.33\textwidth}
\centering
\includegraphics[width=\linewidth]{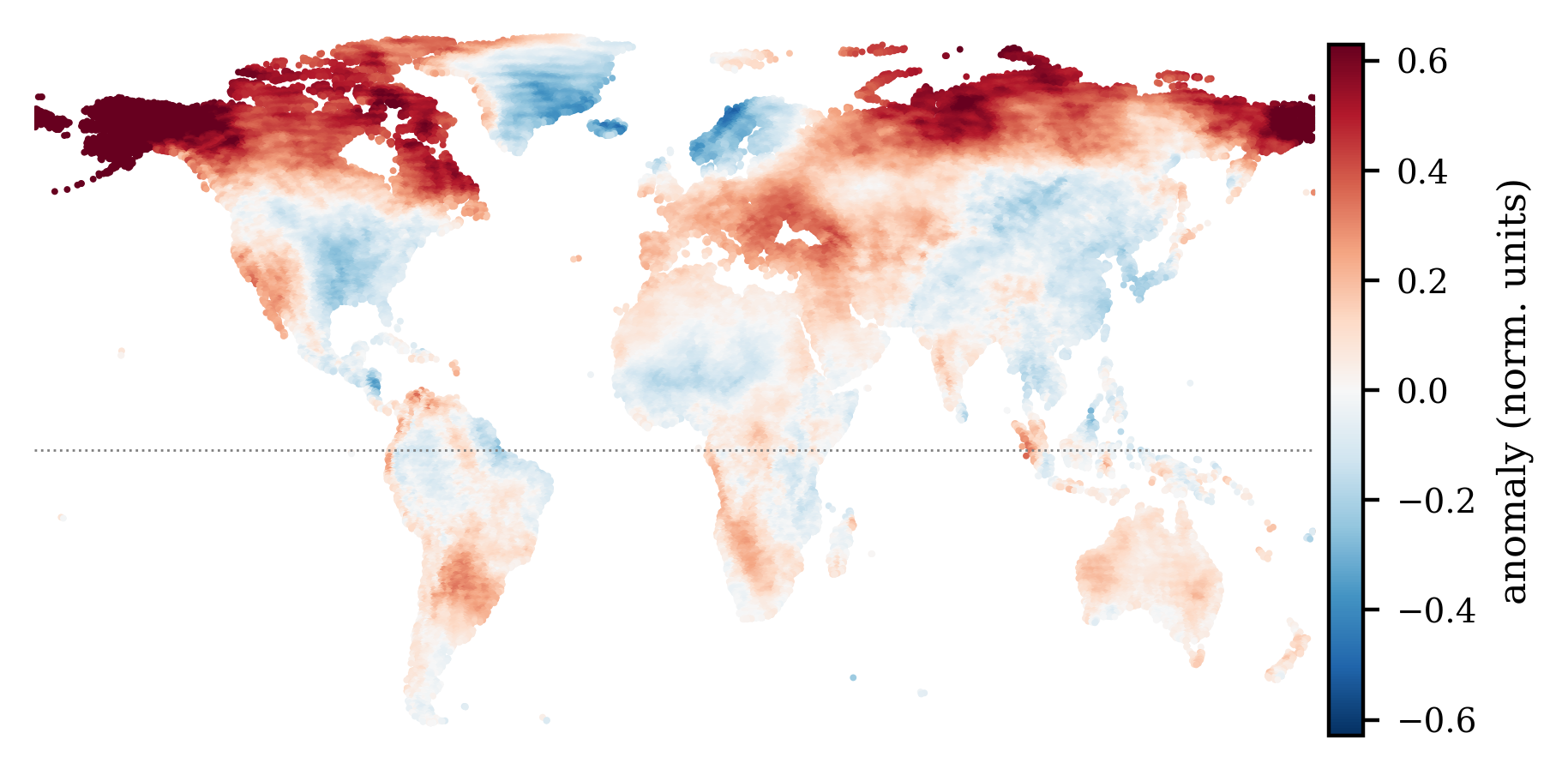}\\[2pt]
{\small Warming trend}
\end{minipage}\hfill
\begin{minipage}[t]{0.33\textwidth}
\centering
\includegraphics[width=\linewidth]{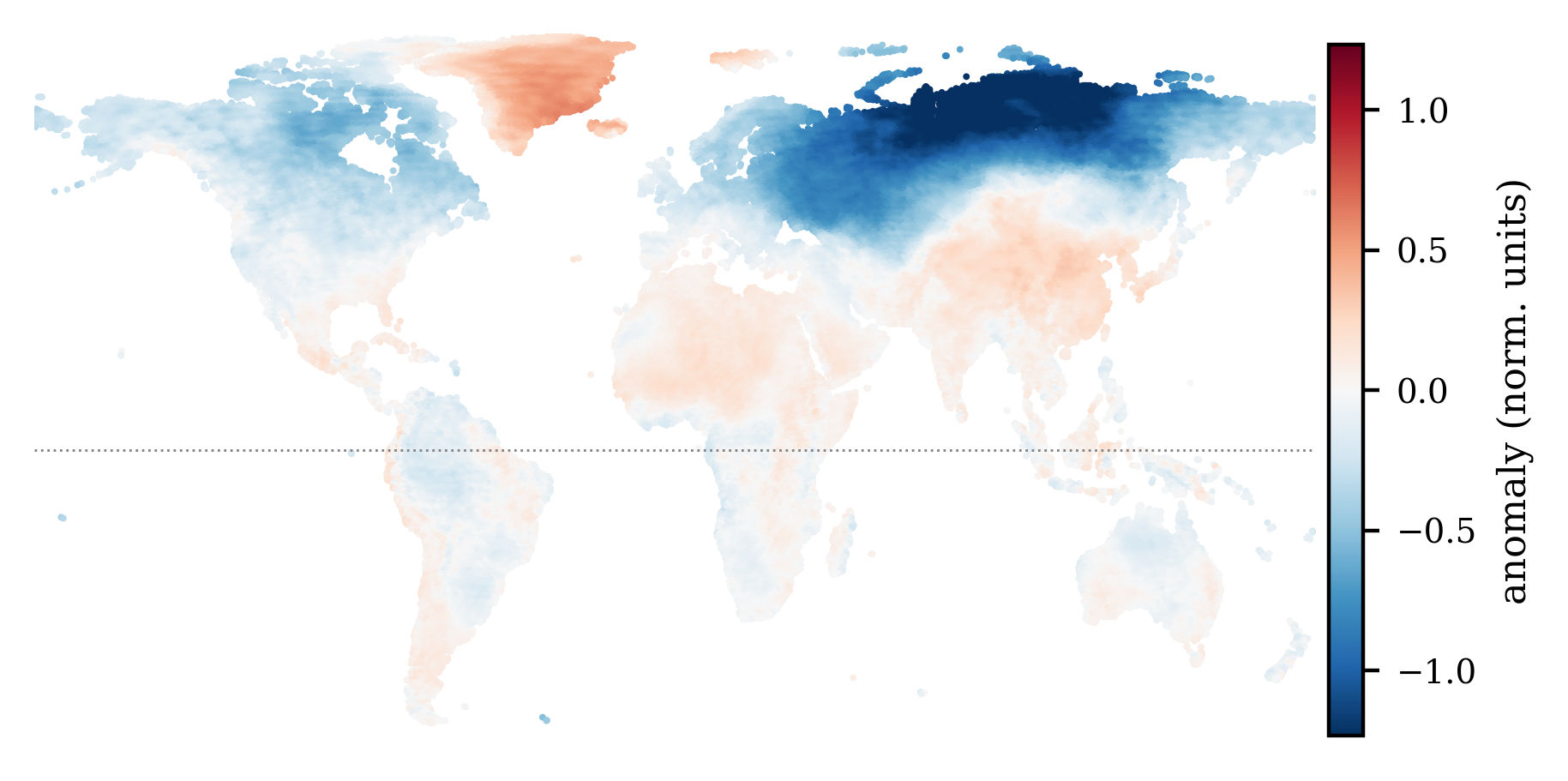}\\[2pt]
{\small El Ni\~no (temperature)}
\end{minipage}\hfill
\begin{minipage}[t]{0.33\textwidth}
\centering
\includegraphics[width=\linewidth]{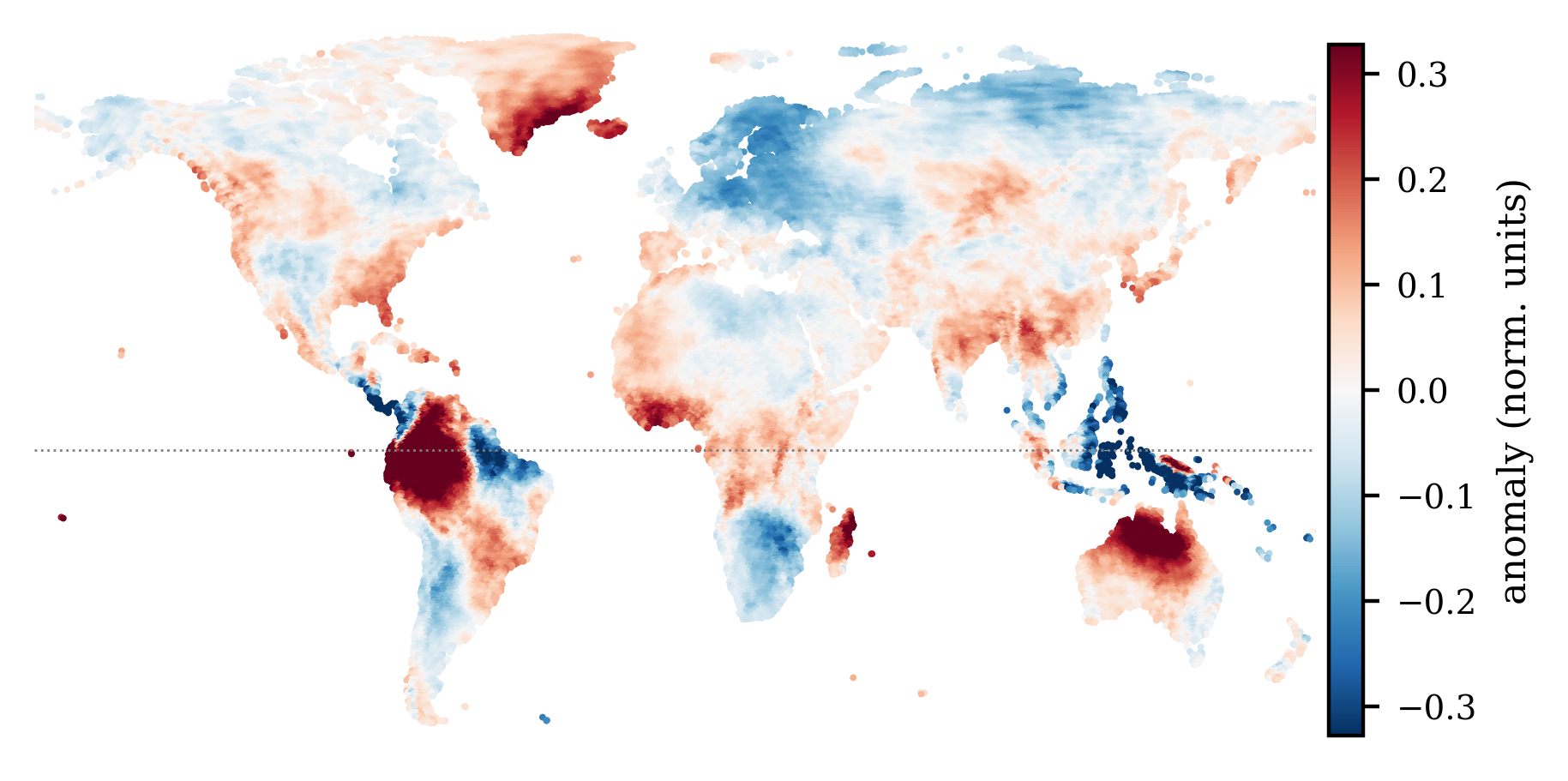}\\[2pt]
{\small El Ni\~no (rainfall)}
\end{minipage}
\caption{\textbf{Index-perturbation maps.} Each panel raises one climate index by $+1\sigma$ (the rest held at their mean) and shows the resulting change in the daily model's mid-January reconstruction over land. Red is warmer or wetter, blue cooler or drier.}
\label{fig:index_response}
\end{figure*}

\subsection{Downstream performance}

\emph{Chronosphere} matches or leads across the spatial and temporal tasks in Table~\ref{tab:main}, and its margin widens under spatial and temporal held-out splits. Figure~\ref{fig:skill_bubble} collapses both axes into one view, where \emph{Chronosphere} sits top-right with the shortest drop from interpolation to extrapolation. Its two variants specialize by resolution. The monthly model trains on CHELSA at ${\sim}1$\,km and leads the spatial tasks. The daily model also trains on ERA5-Land at ${\sim}9$\,km, trading spatial detail for sub-monthly resolution, and leads the within-year tasks.

\paragraph{Spatial tasks.} \emph{Chronosphere} leads every baseline on the three continuous spatial targets under $10^\circ$ holdout: elevation ($0.87$), plant traits ($0.47$), and canopy height ($5.1$\,m). The appendix (Spatial Extrapolation Across Tile Size) traces this as a curve, with \emph{Chronosphere} highest at every separation. On the categorical targets it ties the best on biome and trails only on ecoregion, where CLIMPLICIT edges the iid split ($0.79$ against $0.78$) and STE the extrapolation ($0.32$ against $0.31$), each by one point. \emph{STE}~\cite{mickisch2025ste} is the closest competitor, but \emph{Chronosphere} is the most consistent across targets and the strongest across all experimental regimes.

\paragraph{Within-year tasks.} The ordering follows how each encoder represents time. Encoders without a temporal axis hold one vector per location, so their probes cannot vary through the year and turn negative on snow and streamflow under window extrapolation. The time-aware baselines recover both, but \emph{Chronosphere} leads them on all five within-year targets under both splits. Notably, while CLIMPLICIT and STE train on the same data as the monthly model, Chronosphere reaches $0.85$ and $0.40$ on snow and streamflow extrapolation, while CLIMPLICIT ($0.70$ and $0.12$) and STE ($0.68$ and $0.21$) trail by large margins. This advantage grows on NDVI, snow, and streamflow with finer, daily training data. In addition, as seen in Fig.~\ref{fig:temporal_extrap}, this lead widens under seasonal transfer. As the held-out window moves further from any training season, every encoder loses skill, but \emph{Chronosphere} degrades least and stays above all other encoders.

Beyond these climatology probes, a further experiment uses the frozen embedding as a spatio-temporal prior. We add it as an extra feature to an encoder-free baseline that predicts from past observations, on year-resolved records of fire, streamflow, water storage, dengue, greenness, and snow. Every climate encoder lifts this baseline, and \emph{Chronosphere} lifts it most. The appendix (Encoder as a Geo-Prior) gives the full setup and results.

\subsection{Ablations}

\paragraph{Essential structure.} Table~\ref{tab:ablation_temporal} ablates the daily model on the within-year targets, and two architectural decisions prove necessary. Putting time in the assignment (Eq.~\ref{eq:spacetime}) beats passing the day-of-year to the decoder alone or drifting site embeddings through a seasonal flow. This is most clearly seen on snow extrapolation ($0.85$ against $0.73$ and $0.76$). Removing the local field (Eq.~\ref{eq:field_decoder}) leaves the plain tessellation and the worst row on every target, snow extrapolation falling to $0.79$. An adaptive partition alone does not resolve structure within each region, so the partition and the local field are both essential. Prior encoders supply an adaptive partition or rich local detail but not both, and Table~\ref{tab:main} confirms a global function without a partition trails on nearly every task.

\paragraph{Implementation variants.} The minimal form of each ingredient suffices. Free coefficients give each site direct control of its frequency content, and more elaborate mixing does not help. Gating adds a bottleneck and top-$k$ breaks the smooth blend across experts, both losing ground on streamflow extrapolation ($0.34$ and $0.31$ against $0.40$). Adding capacity changes little with site count saturating by $K{=}8192$. Low frequencies alone fall behind, with most of the signal recovered once higher frequencies are included.

\subsection{Learned temporal structure}

\paragraph{Within-year.} Sampling the location vector through the year and measuring how far it departs from its annual mean, $\lVert z(t)-\bar z\rVert$, maps where the year moves the representation (Fig.~\ref{fig:seasonal}, left). The vector changes most, as expected, where seasonality is strongest, over boreal Siberia and North America, the Sahel, and the South-Asian monsoon, and much less in the perennially wet tropics.

We visualize a few regions in more detail over the year, converting each $256$-dimensional vector to a color through a shared PCA projection to RGB (Fig.~\ref{fig:seasonal}, right). In NE Siberia the vector sits at a cold, dark-blue extreme in January and swings to a warm yellow by July, a temperature-driven cycle along one PCA axis. Below the equator in S-Central Africa it peaks in the opposite season, lush green in the summer of January and red as the dry winter sets in by July, moving along a different, precipitation-driven axis. At the same calendar month the two regions sit at opposite points of their annual loops. We read this as evidence that the embedding encodes both when each place's cycle turns and what drives it.

\paragraph{Inter-annual.} Perturbing single indices reproduces known climate patterns. Greenhouse forcing warms the poles most, and El Ni\~no shifts rainfall over Southeast Asia (Fig.~\ref{fig:index_response}). We show quantitative comparisons in the appendix (Yearly Index Drivers).

\section{Conclusion}
\label{sec:conclusion}

We introduced \emph{Chronosphere}, a spatio-temporal neural field that learns geographic representations, adapting to the data both \emph{where} it spends capacity and \emph{how finely} it captures detail. Its learnable sites live on the spacetime torus $S^2 \times S^1$, and a per-site mixture of local basis-function experts resolves detail within each region. Trained to reconstruct climatology, its embedding matches or leads state-of-the-art location encoders across spatial and temporal probes, with the largest margins under spatial and seasonal transfer. Prior encoders adopt only one of these two ingredients; unifying them drives the gains, and our ablations and baselines show that neither suffices alone.

Several directions follow naturally from here. One is to extend to other spatio-temporal modalities such as satellite or street-level imagery, letting the model allocate resolution depending upon the change in the modality. A more comprehensive model could integrate many of these modalities together. In a similar vein, conditioning on the downstream task could move the model from a generalist representation toward one that learns the frequencies a given task depends on. The climatological model itself would also sharpen with better data. Known regions with sparse observational coverage (deserts, high mountains) rely heavily on interpolation, so those areas may not be as accurate. In addition, the daily variant model is spatially coarser than the monthly model, sacrificing temporal resolution for spatial. Above all we are excited for its use in ecology. Chronosphere lets researchers characterize the environment of almost any place and time, and we hope it becomes a useful representation for the many ecological problems that require climatological context.

\clearpage
\section*{Acknowledgments}
This research used the TGI RAILs advanced compute and data resource, which is
supported by the National Science Foundation (award OAC-2232860) and the Taylor
Geospatial Institute. This work is also supported by the Ann W. and Spencer T.
Olin-Chancellor's Fellowship at Washington University in St. Louis and by the
AI-ACCESS National Research Traineeship, funded by the National Science
Foundation (award DGE-2244165).

\bibliography{aaai2027}

\end{document}